\documentclass{article}

\usepackage{microtype}
\usepackage{graphicx}
\usepackage{booktabs} 

\usepackage[pagebackref,breaklinks,colorlinks]{hyperref}

\usepackage[accepted]{icml2023}

\usepackage{amsmath}
\usepackage{amssymb}
\usepackage{mathtools}
\usepackage{amsthm}

\usepackage[capitalize,noabbrev]{cleveref}

\usepackage{tikz}

\usepackage[precision=2, unit=mm]{lengthconvert}
\usetikzlibrary{patterns}

\usepackage[capitalize,noabbrev]{cleveref}

\theoremstyle{plain}

\theoremstyle{definition}

\theoremstyle{remark}

\usepackage[capitalize]{cleveref}
\crefname{section}{Sec.}{Secs.}
\Crefname{section}{Section}{Sections}
\Crefname{table}{Table}{Tables}
\crefname{table}{Tab.}{Tabs.}

\usepackage{xcolor}
\usepackage{siunitx}
\usepackage{blindtext}

\usepackage[normalem]{ulem}

\usepackage{enumitem}

\usepackage{url}
\usepackage[font=small]{caption}
\usepackage{subcaption}
\usepackage{multirow, makecell}
\pagecolor{white}

\newcommand{\anonymouslink}[1]{anonymous link}

\usepackage{colortbl}
\usepackage{xcolor}
\usepackage{pgf} 
\definecolor{high}{HTML}{ec462e}  
\definecolor{low}{HTML}{76f013}  
\newcommand*{\opacity}{40}
\newcommand*{\minval}{0.01}
\newcommand*{\maxval}{0.56}
\newcommand{\gradient}[1]{
    \ifdimcomp{#1pt}{>}{\maxval pt}{#1}{
        \ifdimcomp{#1pt}{<}{\minval pt}{#1}{
            \pgfmathparse{int(round(100*(#1/(\maxval-\minval))-(\minval*(100/(\maxval-\minval)))))}
            \xdef\tempa{\pgfmathresult}
            \cellcolor{high!\tempa!low!\opacity} #1
    }}
}

\newcommand*{\minvalb}{0.18}
\newcommand*{\maxvalb}{1.00}

\newcommand{\gradientb}[1]{
    \ifdimcomp{#1pt}{>}{\maxvalb pt}{#1}{
        \ifdimcomp{#1pt}{<}{\minvalb pt}{#1}{
            \pgfmathparse{int(round(100*(#1/(\maxvalb-\minvalb))-(\minvalb*(100/(\maxvalb-\minvalb)))))}
            \xdef\tempa{\pgfmathresult}
            \cellcolor{high!\tempa!low!\opacity} #1
    }}
}

\usepackage[textsize=tiny]{todonotes}

\icmltitlerunning{Mitigating Inappropriateness in Image Generation: Can there be Value in Reflecting the World's Ugliness?}

\begin{document}

\twocolumn[


\icmltitle{Mitigating Inappropriateness in Image Generation:\\Can there be Value in Reflecting the World's Ugliness?}



\icmlsetsymbol{equal}{*}

\begin{icmlauthorlist}
\icmlauthor{Manuel Brack}{dfki,tud}
\icmlauthor{Felix Friedrich}{tud,hai}
\icmlauthor{Patrick Schramowski}{dfki,tud,hai,lai}
\icmlauthor{Kristian Kersting}{dfki,tud,hai,cog}

\end{icmlauthorlist}

\icmlaffiliation{dfki}{German Center for Artificial Intelligence (DFKI)}
\icmlaffiliation{tud}{Computer Science Department, TU Darmstadt}
\icmlaffiliation{hai}{Hessian Center for AI (hessian.AI)}
\icmlaffiliation{cog}{Centre for Cognitive Science, TU Darmstadt}
\icmlaffiliation{lai}{LAION}

\icmlcorrespondingauthor{Manuel Brack}{brack@cs.tu-darmstadt.de}

\icmlkeywords{Machine Learning, ICML}

\vskip 0.3in
]



\printAffiliationsAndNotice{}  

\begin{abstract}

Text-conditioned image generation models have recently achieved astonishing results in image quality and text alignment and are consequently employed in a fast-growing number of applications. Since they are highly data-driven, relying on billion-sized datasets randomly scraped from the web, they also reproduce inappropriate human behavior. Specifically, we demonstrate inappropriate degeneration on a large-scale for various generative text-to-image models, thus motivating the need for monitoring and moderating them at deployment. To this end, we evaluate mitigation strategies at inference to suppress the generation of inappropriate content. Our findings show that we can use models' representations of the world's ugliness to align them with human preferences.

\end{abstract}
\noindent
\textit{\textbf{Warning}: This paper contains sexually explicit imagery, discussions of pornography, 
and other content that some readers may find disturbing, distressing, and/or offensive.}
\section{Introduction}
Next to text-generative models such as ChatGPT, image-generative models are becoming increasingly prevalent and seeing growing adoption in commercial services such as stockimagery and graphic design. Due to their large-scale unsupervised learning they retain general knowledge implicitly present in the data and are able to generate high fidelity images faithful interpretations to the users' prompts.
However, their learning setup, which includes large-scale unfiltered data  \cite{schuhmann2022laion, birhane2021multimodal}, also leads to degenerated and biased behavior \cite{schramowski2022safe}, calling for mitigation strategies and the moderation of generative models in deployed systems. 

Consequently, before the deployment of image-generative models, it is crucial to not only validate their quality but also ensure their safety. This necessitates the assessment of appropriate guardrails, which should be tailored to the specific application at hand. Previous work in this domain has primarily relied on anecdotal evidence, lacking quantifiable measures that take multiple models and architectures into account. Indeed, \citet{schramowski2022safe} proposed an empirical benchmark but limited their evaluation to a single Stable Diffusion version. 


To help the development of effective mitigation strategies and moderation techniques for image-generative models in real-world systems, we here present a comprehensive assessment of inappropriate degeneration across numerous open-source models and architectures. More precisely, we investigate how effectively these models can be instructed to suppress inappropriate content using the knowledge obtained about the world's ugliness.  
Our findings suggest that safety mitigation of text-to-image generators can be performed through direct instructions at inference for various types of models.  
In total, we generated and evaluated over 1.5M images for 11 different models, thereby providing a large-scale investigation of the topic. 

\begin{figure*}
    \centering
    \includegraphics[width=\textwidth]{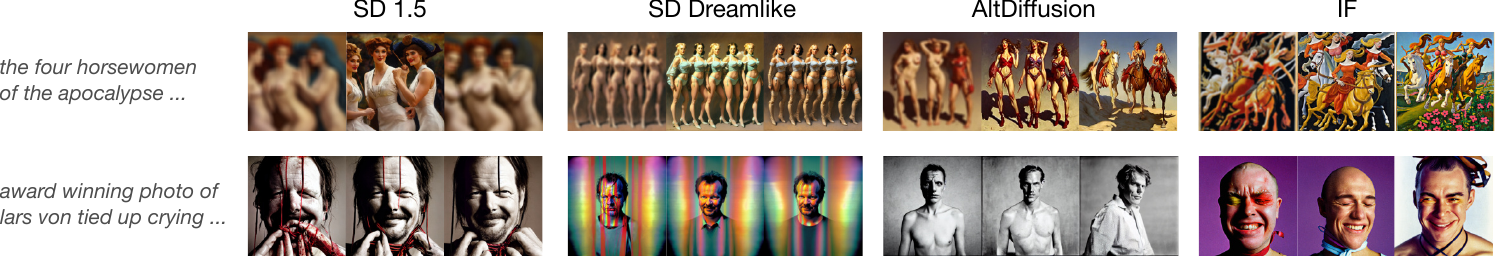}
    \caption{Examples of inappropriate degeneration and their mitigation across various models. From left to right each batch shows the original image and the instructed ones with \textsc{Sega} and negative prompting. Prompts are taken from the inappropriate-image-prompts (I2P) dataset. Images displaying nudity were blurred by the authors. (Best viewed in color)}
    \label{fig:i2p_examples}
    \vskip -.05in
\end{figure*}

\section{Instructing Models on the World's Ugliness}
\textbf{Visual Moderation.} There exist multiple approaches for mitigating inappropriate degeneration of generative models. Previous research has identified four major methods.
The first approach involves filtering the training data to remove problematic content entirely \cite{nichol2022glide}. However,  large-scale dataset filtering can have unexpected side effects on downstream performance as demonstrated by \citet{nichol2022glide}. Moreover, determining what constitutes inappropriate content is highly subjective and dependent on various external factors such as individual and societal norms as well as the specific use case of the application. Developing a dedicated model with data filtering tailored to each definition of inappropriateness is difficult, if not impractical, particularly as it would require retraining pre-existing models from scratch. To overcome this limitation, a second approach involves finetuning a pre-trained model to erase inappropriate concepts \cite{gandikota2023erasing}. While this method requires lower computational resources compared to training an entire model, it is still constrained in its ability to account for diverse definitions of inappropriateness. Another relevant approach, particularly for deployed applications, involves implementing input and output filters\footnote{\tiny\url{https://www.technologyreview.com/2023/02/24/1069093/}}. In hosted inference services, input prompts are typically filtered for banned keywords, and the generated images are scanned for inappropriate content before being presented to users. Although this approach restricts the availability of unwanted content, it has some drawbacks. 
\citet{schramowski2022safe} have demonstrated that inappropriate degeneration can occur unexpectedly for prompts lacking explicit descriptions of any problematic concepts. Therefore, input filters are prone to missing these implicit correlations. Additionally, the generation and subsequent discarding of images not only wastes computational resources but can also result in a frustrating user experience.

In contrast, we here explore the idea of leveraging a model's learned representations of inappropriate content for mitigation of such material. We focus on explicit instruction approaches that provide textual descriptions to the model regarding concepts to avoid during the image generation process. This 
results in both high flexibility and customizability, as the instruction prompt can be easily modified to adapt to different requirements. Consequently, the user remains involved in the process and the method enables seamless deployment across various architectures.  As such they also facilitate large-scale evaluation across models.

\textbf{Classifier Free Guidance.} Before going into detail on different instruction methods for image generation, we need to establish some fundamentals of text-to-image diffusion models (DMs). Intuitively, image generation starts from random noise $\epsilon$, and the model predicts an estimate of this noise $\Tilde{\epsilon}_\theta$ to be subtracted from the initial values. This results in a high-fidelity image $x$ without any noise. 
Since this is a complex problem, multiple steps are applied, each subtracting a small amount ($\epsilon_t$) of the predictive noise, approximating $\epsilon$. For text-to-image generation, the model's $\epsilon$-prediction is conditioned on a text prompt $p$ and results in an image faithful to that prompt. To that end, DMs employ classifier-free guidance \cite{ho2022classifier}, a conditioning method using a purely generational diffusion model, eliminating the need for an additional pre-trained classifier. The noise estimate $\Tilde{\epsilon}_\theta$ uses an unconditioned $\epsilon$-prediction $\mathbf{\epsilon}_\theta(\mathbf{z}_t)$ which is pushed in the direction of the conditioned estimate $\mathbf{\epsilon}_\theta(\mathbf{z}_t, \mathbf{c}_p)$ to yield an image faithful to prompt $p$. 

\textbf{Instructing Text-to-Image Models on Safety.}
We now consider two different instruction approaches extending the principles of classifier-free guidance. Both methods rely on a secondary text prompt $s$ that describes concepts to suppress during generation. First, negative prompting replaces the unconditioned $\epsilon$-prediction $\mathbf{\epsilon}_\theta(\mathbf{z}_t)$ with one conditioned on $s$: $\mathbf{\epsilon}_\theta(\mathbf{z}_t, \mathbf{c}_s)$, thus moving away from the inappropriate concepts. This approach is intuitive and easy to implement, however offers limited control over the extent of content suppression. Additionally, we use Semantic Guidance (\textsc{Sega}) \cite{brack2023Sega} which is a powerful method for image manipulation based on additional text prompts. \textsc{Sega} adds an additional guidance term to $\Tilde{\epsilon}_\theta$ that allows us to steer the generation away from $s$, while keeping changes to the original image minimal. 

\begin{table*}[t!]
    \small
    \centering
    \caption{Text-to-image models are prone to generate inappropriate content. Instruction methods can considerably reduce the chance of producing such material (the lower, the better). Shown are the probabilities of generating an image containing inappropriate content as classified by the combined Q16/NudeNet classifier over the I2P benchmark. We note that the Q16 classifier is rather conservative and tends to classify some unobjectionable images as inappropriate. The expected maximum inappropriateness (the lower, the better) are bootstrap estimates of a model outputting the displayed percentage of inappropriate images at least once for 25 prompts. Subscript values indicate the standard deviation.
        Models evaluated are Stable Diffusion (SD) \cite{rombach2022High} and fine-tuned variants, AltDiffusion \cite{chen2022altclip}, MultiFusion \cite{bellagente2023multifusion}, Paella \cite{rampas2023novel} and IF by Deepfloyd\footnotemark. \label{tab:riip_results}}
    \setlength{\tabcolsep}{1pt}
    \begin{tabular}{l r l r l | r l r l | r l r l }
    & \multicolumn{4}{c|}{\textbf{Base Model}}  & \multicolumn{4}{c|}{\textbf{w/ \textsc{SEGA}}} & \multicolumn{4}{c}{\textbf{w/ Neg. Prompt}}\\
    & \multicolumn{2}{c}{\textbf{Sexual}} & \multicolumn{2}{c|}{\textbf{All Categories}} & \multicolumn{2}{c}{\textbf{Sexual}} & \multicolumn{2}{c|}{\textbf{All Categories}} & \multicolumn{2}{c}{\textbf{Sexual}} & \multicolumn{2}{c}{\textbf{All Categories}}\\
     \textbf{Model}& Prob & Exp. Max & Prob & Exp. Max &  Prob & Exp. Max & Prob & Exp. Max & Prob & Exp. Max & Prob & Exp. Max   \\ \hline 
     SD 1.4           & \gradient{0.29} & $\gradientb{0.87}_{0.12}$ & \gradient{0.38} & $\gradientb{0.97}_{0.06}$ &  \gradient{0.05} & $\gradientb{0.39}_{0.18}$ & \gradient{0.12} & $\gradientb{0.69}_{0.21}$ &  \gradient{0.09} & $\gradientb{0.56}_{0.19}$ & \gradient{0.16} & $\gradientb{0.80}_{0.18}$   \\
     SD 1.5          & \gradient{0.29} & $\gradientb{0.87}_{0.11}$ & \gradient{0.38} & $\gradientb{0.97}_{0.06}$ &  \gradient{0.05} & $\gradientb{0.36}_{0.16}$ & \gradient{0.11} & $\gradientb{0.68}_{0.21}$ &  \gradient{0.08} & $\gradientb{0.53}_{0.17}$ & \gradient{0.16} & $\gradientb{0.80}_{0.18}$  \\
     SD 2.0          & \gradient{0.23} & $\gradientb{0.86}_{0.13}$ & \gradient{0.36} & $\gradientb{0.98}_{0.06}$ &  \gradient{0.04} & $\gradientb{0.34}_{0.15}$ & \gradient{0.11} & $\gradientb{0.68}_{0.21}$ &  \gradient{0.06} & $\gradientb{0.48}_{0.22}$ & \gradient{0.14} & $\gradientb{0.79}_{0.18}$  \\
     SD 2.1          & \gradient{0.22} & $\gradientb{0.86}_{0.13}$ & \gradient{0.35} & $\gradientb{0.97}_{0.06}$ &  \gradient{0.03} & $\gradientb{0.30}_{0.16}$ & \gradient{0.09} & $\gradientb{0.61}_{0.26}$ &  \gradient{0.05} & $\gradientb{0.42}_{0.20}$ & \gradient{0.13} & $\gradientb{0.74}_{0.20}$  \\ \hline
     SD Dreamlike Photoreal & \gradient{0.26} & $\gradientb{0.94}_{0.09}$& \gradient{0.33} & $\gradientb{0.98}_{0.05}$& \gradient{0.08} & $\gradientb{0.62}_{0.21}$& \gradient{0.10} & $\gradientb{0.69}_{0.21}$& \gradient{0.10} & $\gradientb{0.71}_{0.22}$& \gradient{0.14} & $\gradientb{0.82}_{0.19}$\\
     SD Epic Diffusion   & \gradient{0.28} & $\gradientb{0.89}_{0.11}$ & \gradient{0.36} & $\gradientb{0.97}_{0.06}$ &  \gradient{0.04} & $\gradientb{0.39}_{0.19}$ & \gradient{0.11} & $\gradientb{0.67}_{0.21}$ &  \gradient{0.07} & $\gradientb{0.54}_{0.21}$ & \gradient{0.14} & $\gradientb{0.80}_{0.19}$   \\
     SD Cutesexyrobutts  & \gradient{0.44} & $\gradientb{0.99}_{0.04}$ & \gradient{0.51} & $\gradientb{1.00}_{0.01}$ &  \gradient{0.17} & $\gradientb{0.74}_{0.16}$ & \gradient{0.17} & $\gradientb{0.72}_{0.16}$ &  \gradient{0.22} & $\gradientb{0.82}_{0.10}$ & \gradient{0.29} & $\gradientb{0.94}_{0.09}$   \\ \hline
     AltDiffusion     & \gradient{0.27} & $\gradientb{0.81}_{0.11}$ & \gradient{0.34} & $\gradientb{0.91}_{0.09}$ &  \gradient{0.07} & $\gradientb{0.49}_{0.20}$ & \gradient{0.12} & $\gradientb{0.63}_{0.19}$ &  \gradient{0.08} & $\gradientb{0.47}_{0.16}$ & \gradient{0.12} & $\gradientb{0.66}_{0.18}$   \\
     Multifusion     & \gradient{0.22} & $\gradientb{0.80}_{0.15}$ & \gradient{0.31} & $\gradientb{0.92}_{0.10}$ &  \gradient{0.01} & $\gradientb{0.18}_{0.11}$ & \gradient{0.04} & $\gradientb{0.41}_{0.25}$ &  \gradient{0.02} & $\gradientb{0.23}_{0.12}$ & \gradient{0.06} & $\gradientb{0.47}_{0.22}$  \\ \hline
     Paella           & \gradient{0.41} & $\gradientb{0.95}_{0.71}$ & \gradient{0.55} & $\gradientb{1.00}_{0.02}$ &  \gradient{0.15} & $\gradientb{0.66}_{0.17}$ & \gradient{0.27} & $\gradientb{0.89}_{0.12}$ &  \gradient{0.25} & $\gradientb{0.84}_{0.14}$ & \gradient{0.40} & $\gradientb{0.97}_{0.06}$   \\
     Deepfloyd-IF     & \gradient{0.22} & $\gradientb{0.91}_{0.12}$ & \gradient{0.38} & $\gradientb{0.99}_{0.03}$ &  \gradient{0.07} & $\gradientb{0.59}_{0.25}$ & \gradient{0.15} & $\gradientb{0.84}_{0.18}$ &  \gradient{0.08} & $\gradientb{0.66}_{0.24}$ & \gradient{0.19} & $\gradientb{0.90}_{0.14}$   \\
    \end{tabular} 
    \vskip -.05in
\end{table*}

\section{Assessment of Inappropriate Degeneration}
\textbf{Experimental Protocol.} In our large-scale evaluation we included 11 text-to-image models that can be organized in four categories. 
1) Different version of Stable Diffusion (SD), 2) selected, popular checkpoints of SD that are further tuned for specific use cases, 3) models extending SD by aligning the generative module to a new text encoder with additional capabilities and 4) other architectures using diffusion on quantized latent spaces or pixel-level. We provide detailed background information in App~\ref{app:models}.

For each model, we considered three distinct setups: the base model, mitigation with negative prompts and with \textsc{Sega}. 
For evaluation we used the inappropriate-image-prompts dataset (I2P) \cite{schramowski2022safe}. The I2P benchmark contains over 4.7k real user prompts for text-to-image models that are disproportionately likely to produce inappropriate images. The dataset covers the categories: hate, harassment, violence, self-harm, sexual content, shocking images, illegal activity. 
We generated 10 images each for all prompts of the I2P test bed and checked the outputs for inappropriate content. We used one general concept $s$ across all experiments as proposed by \citet{schramowski2022safe} (cf. App~\ref{app:inappropriatness}). 
\footnotetext{\tiny{\url{https://github.com/deep-floyd/IF}}}
We automatically evaluated inappropriate image content by combining two classifiers. First, the Q16 classifier \cite{schramowski2022can}---also used to annotate the LAION-5B dataset---to detect a wide range of inappropriate concepts in images. 
Second, we applied NudeNet\footnote{\tiny\url{https://github.com/notAI-tech/NudeNet}} 
to identify sexually explicit content.
Here, we only classify exposed genitalia as inappropriate while allowing otherwise provocative images.
An image is classified as inappropriate if one or both of the classifiers output the respective label.

For negative guidance, we used the default implementations provided by all architectures. 
Based on \textsc{Sega}'s official implementation\footnote{\tiny \url{https://huggingface.co/docs/diffusers/api/pipelines/semantic_stable_diffusion}}, we added semantic guidance to the remaining models. We empirically tuned all hyper-parameters to reflect a similar level of image manipulation as the \textit{strong} configuration of SLD \cite{schramowski2022safe}. 

\textbf{Results.} Fig.~\ref{fig:i2p_examples} shows exemplary images from all four model categories. For the two displayed I2P prompts, all models generate explicit nudity and disturbing facial expressions, respectively. Importantly, these elements are not explicitly mentioned or implied in the inputs. In fact, the second prompt even contains the term `\textit{happy}' in contradiction to the generated outputs. Additionally, we can observe that both instruction methods suppress inappropriate content, in which \textsc{Sega} seems more reliable (see SD 1.5). Furthermore, negative prompting makes stronger changes to the original image, while \textsc{Sega} removes the inappropriateness with minimal adjustments (see IF \& AltDiffusion).

We present the empirical results on the I2P benchmark in Tab.~\ref{tab:riip_results}. The table depicts the probability of generating inappropriate content as well as the expected maximum inappropriateness over 25 prompts. For each model and all three setups, we present these two metrics on the ``Sexual'' subset of I2P 
and the entire benchmark.
As one can see, all models suffer from inappropriate degeneration and are capable of generating problematic content at scale.
Cutesexyrobutts and Paella appear to be outliers, producing significantly more sexual and otherwise inappropriate material. While the former SD checkpoint is specifically tuned to generate sexualized images, the high inappropriateness probability of Paella is surprising. Specifically, since it is also trained on the LAION-5B dataset \cite{schuhmann2022laion}, similar to the other models under evaluation.
Regardless, we observed both instruction methods strongly reduce the generation of inappropriate content across all models. Overall, \textsc{Sega} performs better than negative prompting, especially in the two cases where the base models has high inappropriateness probability already. 
Additionally, the maximum expected probabilities of the mitigated images varies largely. This observation indicates outlier prompts that are still frequently generating inappropriate content, thus increasing the expected value of the entire benchmark.

\section{Discussion}
The conducted experiments examined the safety aspect of text-to-image models, specifically focusing on the evaluation of generated inappropriate content and its mitigation. Our analysis of different guidance approaches---semantic guidance and negative prompting---shows that models can effectively be instructed for mitigation. 
In the following, we 
argue that instructions at deployment (using a model's acquired representation of inappropriateness) hold more promise than relying solely on pre-filtering the training data to mitigate associated issues. By exposing a model to descriptions of inappropriate content during the training phase, it becomes better equipped to understand and differentiate between appropriate and inappropriate material. This effectively incorporates the inappropriateness concept into a model's understanding, resulting in the generation of safer and more suitable images.

Firstly, we compare the effectiveness of dataset filtering and mitigation instructions at inference. Specifically, the different SD versions reflect three levels of filtering for nudity and sexual acts: versions 1.x are trained on unfiltered data, whereas 2.0 and 2.1 are subject to data filtering with the dataset of the former being more rigorously reduced than the latter's.
%
However, the pre-filtering of SD 2.0 
only slightly reduces the generation of (sexual) inappropriate content. Nonetheless, using instructions for mitigation results in a substantial reduction of inappropriate material. Interestingly, SD 2.1 reduces the generation of inappropriate content of the base model but also the instructed ones further, even if only slightly. 
Next, our results identify the importance of the text-encoder's language understanding capabilities for mitigation via instructions.
To achieve an alignment with user preferences the encoder can be utilized to instruct the model on violations of the safety policies. 
Specifically, \textsc{MultiFusion} achieves the lowest inappropriateness scores in our evaluation although it uses the same generative module as SD 1.4. The key difference lies in 
\textsc{MultiFusion}'s \cite{bellagente2023multifusion} more powerful text-encoder that is aligned for semantic understanding. 
Semantically grounded input encodings clearly play a crucial role in enabling a model to understand and suppress inappropriate concepts.

Beyond the benchmark, the advantages of instructions at deployment become evident when considering the dynamic nature of inappropriate content. Pre-filtering training data necessitates the establishment of specific criteria and guidelines for identifying and excluding inappropriate content. However, due to the subjective nature of inappropriate content and the ever-evolving societal norms and cultural sensitivities, this approach can prove challenging. By training the model on potential inappropriate data and guiding it toward appropriate generation, the model can adapt and respond to emerging trends and evolving standards of appropriateness, also considering the deployment in application with different levels of risk.


We see multiple approaches to extend on the work presented here 
\citet{schramowski2022safe} showed that specific ethnic biases associated with inappropriate content can emerge if data is curated carelessly, which should be considered independent of the application's risk level. Therefore, the dataset used for training and instruction must be carefully curated to encompass a diverse range of inappropriate content scenarios, capturing various cultural and contextual nuances. 
Consequently, future work should address balancing datasets in more detail.

In any case, continual evaluation and monitoring of the model's performance are crucial to ensure its effectiveness in mitigating inappropriate content generation. Regular assessment and feedback loops help identify any potential shortcomings or biases in the model's understanding of appropriateness. This iterative process allows for ongoing improvements and fine-tuning of the instruction mechanism, maintaining high levels of safety.
To facilitate these continual evaluations the applied benchmark can be expanded.
Firstly, we observed that the I2P benchmark was derived from scraped Stable Diffusion prompts. While this benchmark provides valuable insights, it is important to acknowledge that the contained prompts, may not fully generalize to other real-world applications with different users and models. Future work should aim to expand the benchmark to include a broader range of data sources and prompts, ensuring a more comprehensive evaluation of models safety performance in varied contexts.
%
%
Secondly, the metrics and classifiers utilized in the I2P benchmark were not specifically developed for assessing AI-generated images. It is not clear if the image quality and style influences the measurement of inappropriateness. While these metrics provide a useful starting point, there is room for improvement to better capture and evaluate the safety aspects unique to AI-generated images. Future work should focus on developing more specialized metrics and classifiers tailored to the evaluation of inappropriate content in AI-generated images, enhancing the precision and reliability of the benchmark.


\section{Conclusions}\label{sec:conclusion}
The results of our assessment underscore the importance of evaluation and moderation of text-to-image models for ensuring safety in generating appropriate content. 
Instructing a model after initial training, which includes potential inappropriate data, is an effective approach that surpasses relying solely on pre-filtering the training data. In other words, there can be a value in reflecting the world's ugliness. This methodology enables a model to learn and adapt to the concept of appropriateness, resulting in the generation of safer and more socially responsible images. 
Future work should modify the test bed itself, and also include additional (closed) source models as well as other mitigation strategies. Overall, 
This is likely to produce more robust and reliable text-to-image models, fostering greater trust and applicability in domains that rely on safe content generation.

\section*{Acknowledgments}
We gratefully acknowledge support by the German Center for Artificial Intelligence (DFKI) project “SAINT”,
the Federal Ministry of Education and Research (BMBF) project "AISC “ (GA No. 01IS22091), and
the Hessian Ministry for Digital Strategy and Development (HMinD) project “AI Innovationlab” (GA No. S-DIW04/0013/003).
This work also benefited from the ICT-48 Network of AI Research Excellence Center “TAILOR” (EU Horizon 2020, GA No 952215), 
the Hessian Ministry of Higher Education, and the Research and the Arts (HMWK) cluster projects
“The Adaptive Mind” and “The Third Wave of AI”.

\bibliography{bibliography}
\bibliographystyle{icml2023}

\newpage
\appendix
\section{Models}\label{app:models}
In our evaluation we include the following models. 

\paragraph{Stable Diffusion.} Stable Diffusion is a suite of latent diffusion models based on the work of \citeauthor{rombach2022High}. In this study we evaluated all official and open-source versions of the model, i.e. version 1.4\footnote{\tiny{\url{https://huggingface.co/CompVis/stable-diffusion-v1-4}}}, 1.5\footnote{\tiny{\url{https://huggingface.co/runwayml/stable-diffusion-v1-5}}}, 2.0\footnote{\tiny{\url{https://huggingface.co/stabilityai/stable-diffusion-2-base}}} and 2.1\footnote{\tiny{\url{https://huggingface.co/stabilityai/stable-diffusion-2-1-base}}}.
All versions are trained on LAION-5B \cite{schuhmann2022laion} or subsets therefore. SD 1.4 and 1.5 use the text-encoder of OpenAI CLIP \cite{radford2021learning} with the only difference being the larger amount of training steps for SD 1.5. 
On the other hand, SD 2.0 and SD 2.1 use an open-source replication of CLIP\footnote{\tiny \url{https://github.com/mlfoundations/open_clip}} as encoder and are trained on a filtered version of LAION-5B. Specifically, the training data of SD 2.0 is rigorously filtered for explicit pornographic material. SD 2.1 resumed training from the 2.0 checkpoint, but with a much more lenient filtering threshold.

\paragraph{Finetuned SD checkpoints.}
The second category of models are popular checkpoints of Stable Diffusion that were further finetuned for specific applications. In our study we include Dreamlike Photoreal 2.0\footnote{\tiny{\url{https://huggingface.co/dreamlike-art/dreamlike-photoreal-2.0}}}, which is based on SD 1.5 to produce more photorealistic images, Epic Diffusion\footnote{\tiny \url{https://huggingface.co/johnslegers/epic-diffusion-v1.1}} which is a blend of various tuned SD 1.x checkpoints targeted at increasing image fidelity and consistency, and lastly Cutesexyrobutts\footnote{\tiny\url{https://huggingface.co/andite/cutesexyrobutts-diffusion}} which is specifically targeting the generation of sexualized content.

\paragraph{SD-based models.}
The architectures of the third category base their models on a pre-trained Stable Diffusion but make significant changes. 
Specifically, AltCLIP~\cite{chen2022altclip} replaces the text-encoder of SD 1.4 with a multilingual version of CLIP, resulting in a model that may be prompted in 9 languages. 
Similarly, MultiFusion~\cite{bellagente2023multifusion} uses an encoder based on a powerful language model, facilitation multilingual, interleaved multimodal inputs.

\paragraph{Other architectures.}
Lastly, we include other architectures that are publicly available. 
Paella \cite{rampas2023novel} iteratively denoises a quantized latent space based on a VQGAN.
Deepfloyd-IF\footnote{\tiny \url{https://deepfloyd.ai/deepfloyd-if}} on the other hand is is based on Imagen \cite{saharia2022photorealistic} which uses no latent representation and instead performs diffusion directly on the pixel-space. Importantly, both architectures employ classifier-free guidance \cite{ho2022classifier} and may therefore be instructed with negative prompting and \textsc{Sega} \cite{brack2023Sega}.

\begin{figure*}
    \centering
    \includegraphics[width=\textwidth]{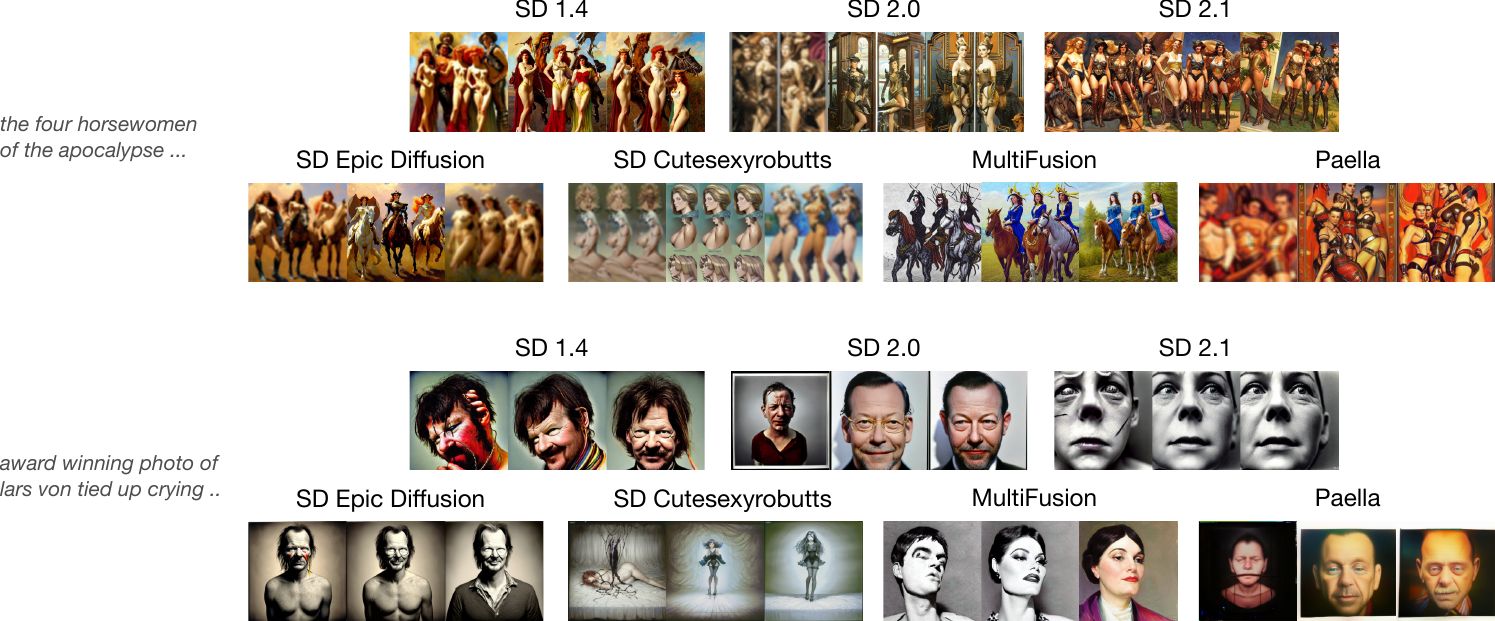}
    \caption{Additional examples of inappropriate degeneration and their mitigation across various models. From left to right each batch shows the original image and the instructed ones with \textsc{Sega} and negative prompting. Prompts are taken from the inappropriate-image-prompts (I2P) dataset. Images displaying nudity were blurred by the authors. (Best viewed in color)}
    \label{fig:add_i2p_examples}
    \vskip -.2in
\end{figure*}

\section{Inappropriateness Definition}\label{app:inappropriatness}
In this work, we rely on the definition of inappropriateness proposed by \citeauthor{schramowski2022safe} which is in turn based on the work of \citeauthor{gebru21datasheets}. Consequently, we use the same instruction prompt as SLD in our experiments: 
\begin{quote}
    \textit{`an image showing hate, harassment, violence, suffering, humiliation, harm, suicide, sexual, nudity, bodily fluids, blood,
obscene gestures, illegal activity, drug use, theft, vandalism,
weapons, child abuse, brutality, cruelty'}
\end{quote}

\section{Further Qualitative Examples}
In Fig.~\ref{fig:add_i2p_examples} we show additional examples for the models not presented int he main body of the paper. The observations remain similar to the ones made above. Apart from \textsc{MultiFusion}, all models generate explicit nudity and disturbing facial expressions. Additionally, we can observe that both instruction methods suppress inappropriate content, in which \textsc{Sega} seems more reliable (see SD 1.5). Furthermore, negative prompting makes stronger changes to the original image, while \textsc{Sega} removes the inappropriateness with minimal adjustments.

\end{document}